\documentclass[pdflatex,sn-mathphys-num]{sn-jnl}% Math and Physical Sciences Numbered Reference Style
\usepackage{graphicx}%
\usepackage{multirow}%
\usepackage{amsmath,amssymb,amsfonts}%
\usepackage{amsthm}%
\usepackage{mathrsfs}%
\usepackage[title]{appendix}%
\usepackage{xcolor}%
\usepackage{textcomp}%
\usepackage{manyfoot}%
\usepackage{booktabs}%
\usepackage{algorithm}%
\usepackage{algorithmicx}%
\usepackage{algpseudocode}%
\usepackage{listings}%
\usepackage{hyperref}
\theoremstyle{thmstyleone}%
\theoremstyle{thmstyletwo}%

\theoremstyle{thmstylethree}%

\begin{document}

\title[Article Title]{Classical Hardware Acceleration of Quantum Autoencoders for Real-Time Anomaly Detection in Collider Experiments}

\author*[1]{\fnm{Ivan} \sur{Ge}}\email{ivange@stanford.edu}

\author*[2]{\fnm{Sagar} \sur{Addepalli}}\email{sagar@slac.stanford.edu}

\author[2]{\fnm{Abhilasha} \sur{Dave}}\email{adave@slac.stanford.edu}

\author[2]{\fnm{Julia} \sur{Gonski}}\email{jgonski@slac.stanford.edu}

\affil[1]{\orgdiv{Department of Physics}, \orgname{Stanford University}, \orgaddress{\street{450 Jane Stanford Way}, \city{Stanford}, \postcode{94305}, \state{CA}, \country{USA}}}

\affil[2]{\orgname{SLAC National Accelerator Laboratory}, \orgaddress{\street{2575 Sand Hill Road}, \city{Menlo Park}, \postcode{94025}, \state{CA}, \country{USA}}}

\abstract{Quantum machine learning (QML) algorithms in high energy physics (HEP) can efficiently represent and leverage long-range, high-order correlations in high-dimensional collider data, potentially with fewer parameters and favorable scaling relative to classical models. Deployment of QML in real-time collider applications such as trigger systems requires the ability to emulate and compile quantum circuits classically, then synthesize the resulting quantum gates onto low-latency hardware accelerators, namely field-programmable gate arrays (FPGAs). We present a study of variational quantum autoencoder models for real-time anomaly detection triggers in modern collider experiments. The models achieve performance comparable to state-of-the-art classical approaches and, after FPGA synthesis, satisfy resource usage and timing constraints consistent with trigger applications in future colliders. This work provides one of the first FPGA implementations of QML models for HEP triggers, enabling higher-capability models in today's classical data acquisition pipelines while advancing quantum readiness of collider experiment infrastructure.}


\maketitle

\section{Introduction}\label{sec:intro}

Research in high energy physics (HEP) aims to test and extend the Standard Model description of fundamental particles and forces. 
HEP experiments operate at the frontier of data throughput and complexity; experiments such as the ATLAS detector at the Large Hadron Collider (LHC) produce tens of terabytes per second~\cite{Aad_2024} of digitized signals produced by outgoing particles from hadron collisions.
Such modern collider experiments record millions to billions of collisions, each containing thousands of low-level detector measurements, making collider physics an ideal setting for machine learning (ML)-driven discovery.
Multi-layered trigger and data acquisition systems~\cite{2137107, Collaboration:2759072} at collider experiments are essential reduce the data rates to values appropriate for storage and analysis.

Classical ML methods, such as decision trees and neural networks, are ubiquitous in HEP research and continue to deliver on key physics drivers such as precision Higgs measurements and beyond the Standard Model (BSM) discovery potential~\cite{hepmllivingreview}.
Quantum machine learning (QML)~\cite{Biamonte_2017} offers a promising alternative approach~\cite{doi:10.1126/science.abn7293} to classical ML for such data intensive tasks where extremely rare signals manifest as deviations in subtle high-dimensional long-range correlations between features.
In QML, learning models are typically implemented as gate-based quantum circuits: a feature map (a sequence of parameterized single- and multi-qubit gates) embeds classical inputs into a quantum state, and an ansatz composed of trainable quantum gates (for example, entangling operations to couple qubits and build nontrivial correlations) forms the hypothesis class.
Measurements of the final state define the model outputs. 
Applications to HEP of quantum, quantum-inspired, and hybrid classical-quantum algorithms span a variety of use cases, including simulation~\cite{PhysRevLett.126.062001} and specialized event selection~\cite{Felser:2020mka}.

Beyond the high learning potential, QML algorithms offer efficient implementations requiring significantly fewer learnable parameters and potentially fewer operations than using classical algorithms for similar tasks. 
These benefits favorably position QML algorithms for running in classical hardware accelerators, such as the field programmable gate arrays (FPGAs) comprising the ATLAS Level-1 trigger which must effectively filter the initial 40 MHz event rate. 
FPGA-based ML also has applications in quantum systems controls~\cite{linne2026squadsmartquantumdetection}. 
The ability to simulate quantum algorithms on FPGAs also opens the door to testing QML applications in sub-millisecond or otherwise constrained ``edge" scenarios where a GPU is not suitable. 
Emulating quantum circuits on FPGAs is a growing field, with recent developments in gate, circuit, and algorithmic implementations ~\cite{1032648.1033377, Conti_2024, 8115369, Hong_2022bie, 10.1109/ISVLSI.2008.43,vanduy2025hpqeascalablehighperformancequantum}. Recent toolchains have also been developed that compile circuits into RTL emulator architectures or HLS-based state-vector simulators ~\cite{volpe2025aequam, bennakhi2025q2sv}.

To assess the feasibility of QML for hardware-accelerated real-time usage in collider physics, this study considers trigger-level anomaly detection (AD).
Here, AD leverages ML to learn a background distribution (typically directly from data) in order to identify elements that are out-of-distribution or non-background-like. 
This paradigm is particularly valuable in HEP because the exact nature of BSM physics is unknown, making it difficult and inefficient to solely pursue dedicated searches for specific signal hypotheses~\cite{Belis_2024}. 
AD is being explored broadly at the LHC, with multiple studies and searches performed at the ATLAS~\cite{HDBS-2018-59,HDBS-2019-23,EXOT-2021-19,EXOT-2022-07,HMBS-2024-34} and CMS experiments~\cite{cmsad}.
Incorporating AD into the trigger offers a path to fundamentally new BSM phase space~\cite{Govorkova_2022}, but must be operable under strict resource and latency budgets with robustness to changing detector and beam conditions. 

The variational autoencoder (VAE)~\cite{kingma2022autoencodingvariationalbayes} is a standard workhorse for ML-based AD, learning a background distribution by minimizing reconstruction error after encoding data through lossy compression. 
Quantum autoencoders (QAEs) are variational QML algorithms which implement the encoder/decoder as parameterized quantum circuits acting on qubits and can be trained analogously from quantum measurements~\cite{ludmir2025quorumzerotrainingunsupervisedanomaly}. 
QML models for HEP have been explored in several forms in AD with demonstrated sensitivity to anomalous LHC events   ~\cite{Ngairangbam_2022, duffy2024unsupervisedbeyondstandardmodeleventdiscovery}. 
Furthermore, quantum-inspired models such as tensor networks (TNs) have been considered  AD~\cite{puljak2025tensornetworkanomalydetection} as well as FPGA-accelerated TNs for AD~\cite{addepalli2026hardwareawaretensornetworksrealtime} and jet tagging~\cite{coppi2026tensornetworkmodelslowlatency}.

In this work, we demonstrate quantum-autoencoder-based AD that can be emulated on FPGAs for collider trigger systems without sacrificing sensitivity to rare BSM signatures. 
We consider both a hybrid quantum–classical VAE, in which quantum circuits are used as part of the encoder/decoder within an otherwise classical latent-variable model, and a fully quantum autoencoder (QAE) architecture. 
The investigation has two components. 
First, we evaluate physics performance by testing AD capability across several BSM benchmark signal models chosen to be highly disparate in their detector-level signatures. 
Second, we evaluate real-time feasibility by synthesizing the trained models for FPGA deployment. 
Using agentic AI–based optimization workflows, we assess resource utilization and end-to-end inference latency and benchmark both against the requirements anticipated for trigger systems in upcoming collider runs.

%--------------------------------------------------------------------------------------------------------------
\section{Methodology}

%--------
ML models are developed using a dataset of simulated proton-proton collision events at the LHC~\cite{govorkova2021lhcphysicsdatasetunsupervised}, pre-filtered to require the presence of at least one energetic electron or muon. 
This dataset includes five simulated processes: a background consisting of multijet events arising from quantum chromodynamics (QCD), and four beyond the Standard Model signals: a neutral scalar boson $A$ decaying via two $Z$ bosons to a four-lepton final state, $A\to4\ell$; a leptoquark (LQ) decaying to a $b$ quark and $\tau$ lepton, $LQ\to b\tau$; acharged scalar boson $h^\pm$ decaying to a $\tau$ lepton and neutrino $\nu$, $h^\pm\to\tau\nu$; and a neutral scalar boson $h^0$ decaying to two $\tau$ leptons, $h^0\to\tau\tau$.

These events are represented by classical input features which are encoded into a quantum state by single-qubit rotation gates $R_X(\theta)$, $R_Y(\theta)$, and $R_Z(\theta)$, which are parameterized quantum operations that rotate a qubit's state on the Bloch sphere by an angle $\theta$ about the $x$-, $y$-, and $z$-axes, respectively. They are defined as
\begin{equation}
R_\alpha(\theta) = \exp\!\left(-i\frac{\theta}{2}\sigma_\alpha\right), \qquad \alpha \in \{x,y,z\},
\end{equation}
where $\sigma_\alpha$ are the Pauli operators. In variational quantum algorithms, these gates typically serve as trainable parameters (and/or data-encoding ``feature map'' parameters), controlling the amplitudes and relative phases of the quantum state and thereby determining the model's expressive power.

%--------
\subsection{Particle Embedding}
\label{sec:embedding}

Each event is represented by 56 real-valued features describing 19 reconstructed objects: missing transverse energy ($E_\mathrm{T}^\mathrm{miss}$), the four
leading electrons, the four leading muons, and the ten leading jets. 
Following the one-particle-one-qubit convention of Ref.~\cite{Bal_2025}, each
reconstructed object is assigned to a single qubit, with its kinematic features encoded through single-qubit rotations. For non-$E_\mathrm{T}^\mathrm{miss}$ particles, we use 
\begin{equation}
    U(p_\mathrm{T},\eta,\phi)
    =
    R_X(\eta)\,
    R_Y(p_\mathrm{T})\,
    R_Z(\phi),
\end{equation}
while the $E_\mathrm{T}^\mathrm{miss}$ qubit omits the $R_X(\eta)$ rotation because
$E_\mathrm{T}^\mathrm{miss}$ is measured only in the transverse plane. We map the raw kinematic features to rotation angles using
\begin{align}
\theta_{p_{\mathrm{T}}}
&= \pi\,\mathrm{clip}\!\left(\frac{\log(1+p_{\mathrm{T}})}{8},\,0,\,1\right), \\
\theta_{\eta}
&= \pi\,\mathrm{clip}\!\left(\frac{\eta}{3},\,-1,\,1\right), \\
\theta_{\phi}
&= \mathrm{clip}\!\left(\phi,\,-\pi,\,\pi\right).
\end{align}
The resulting angles satisfy
\(\theta_{p_{\mathrm{T}}}\in[0,\pi]\) and
\(\theta_{\eta},\theta_{\phi}\in[-\pi,\pi]\) such as to avoid undesirable wraparound effects.
To make the encoding compatible with low-latency FPGA emulation, the full
19-qubit event register is decomposed into smaller particle-type sub-blocks.
The MET, electron, and muon systems form 1-, 4-, and 4-qubit blocks,
respectively. The ten jets present the main hardware challenge: fully
entangling all ten jet qubits would require storing and evolving a
$2^{10}$-amplitude state vector, which is incompatible with the real-time
resource constraints targeted in this work. 
We therefore split the jets into
two independently entangled 5-qubit sub-blocks, reducing the maximum simulated
jet register from 1024 amplitudes to two 32-amplitude registers.
The jet bisection is based on pairwise quantum mutual information ~\cite{Acharya2022QubitSeriation, addepalli2026hardwareawaretensornetworksrealtime}, which recovers the original $p_T$ ordering, so the sub-blocks are simply the five leading and five subleading jets.

%--------
\subsection{Models} 

\subsubsection{Quantum-Classical Hybrid VAE} 
As an initial exploration, we consider a quantum-classical hybrid version of the VAE architecture used in Ref.\cite{Govorkova_2022}. 
In this model, the first dense layer of the encoder is replaced by a block-structured parameterized quantum circuit (PQC). 
Within each block, particle features are encoded using single-qubit rotations, with each particle mapped through $R_X(\eta_k)\,R_Y(p_{\mathrm{T},k})\,R_Z(\phi_k)$, followed by a nearest-neighbor CNOT chain.
Three layers of trainable single-qubit rotations and CNOT entanglers are then applied. 
Each block measures a fixed set of Pauli observables, which are concatenated into a 32-dimensional quantum feature vector. Specifically, the $E_\mathrm{T}^\mathrm{miss}$ block contributes $\langle Z\rangle$, $\langle X\rangle$, and $\langle Y\rangle$; each lepton block contributes four $\langle Z\rangle$ and four $\langle X\rangle$ measurements; jet$_0$ contributes five $\langle Z\rangle$ measurements along with $\langle Z_3 Z_4\rangle$ and $\langle X_3 X_4\rangle$ correlators; and jet$_1$ contributes five $\langle Z\rangle$ measurements and one $\langle Z_3 Z_4\rangle$ correlator. 
The two-wire correlator measurements on the jet sub-blocks are included to recover cross-jet information that would otherwise be lost when splitting the ten jet inputs into two smaller quantum registers. 
The resulting 32-dimensional quantum feature vector is passed to a classical dense layer with ReLU activation, mapping $32 \to 16$. Two parallel linear layers then produce the mean and log-variance of a 3-dimensional Gaussian latent space, each mapping $16 \to 3$. 

\begin{figure} 
\centering 
\includegraphics[width=0.9\textwidth]{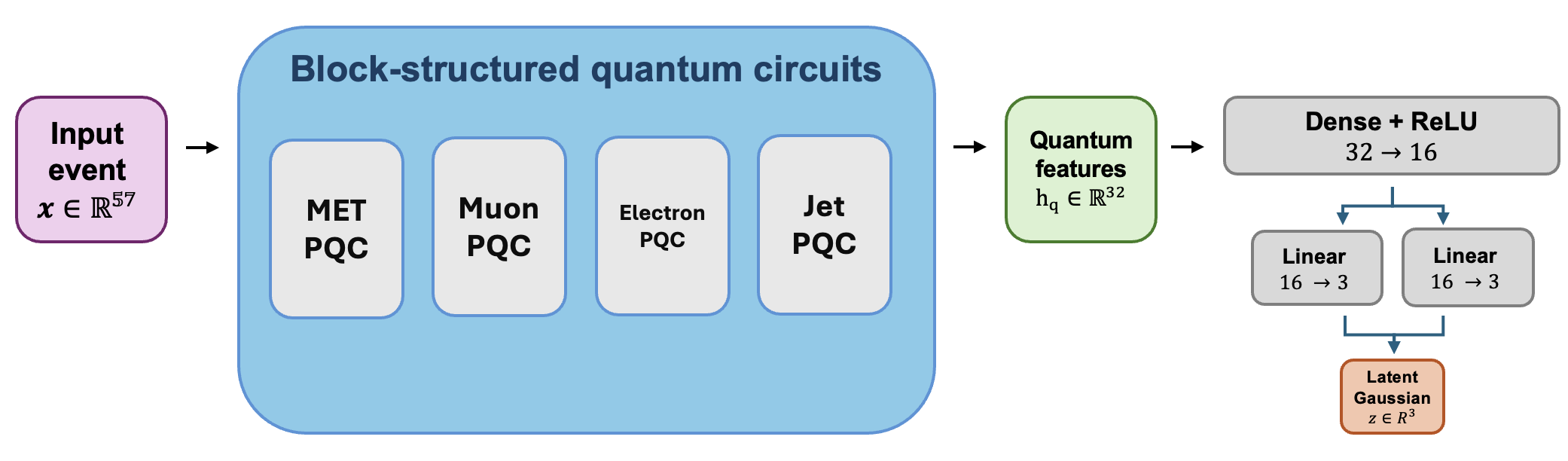} 
\caption{Architecture of the quantum-classical hybrid VAE encoder. The first classical encoder layer is replaced by block-structured PQCs acting on particle-type sub-blocks. The measured quantum observables are concatenated into a 32-dimensional representation, which is mapped to a 3-dimensional Gaussian latent space by classical dense layers. \label{fig:arch_hybrid}} 
\end{figure}

\begin{figure} 
\centering 
\includegraphics[width=0.95\textwidth]{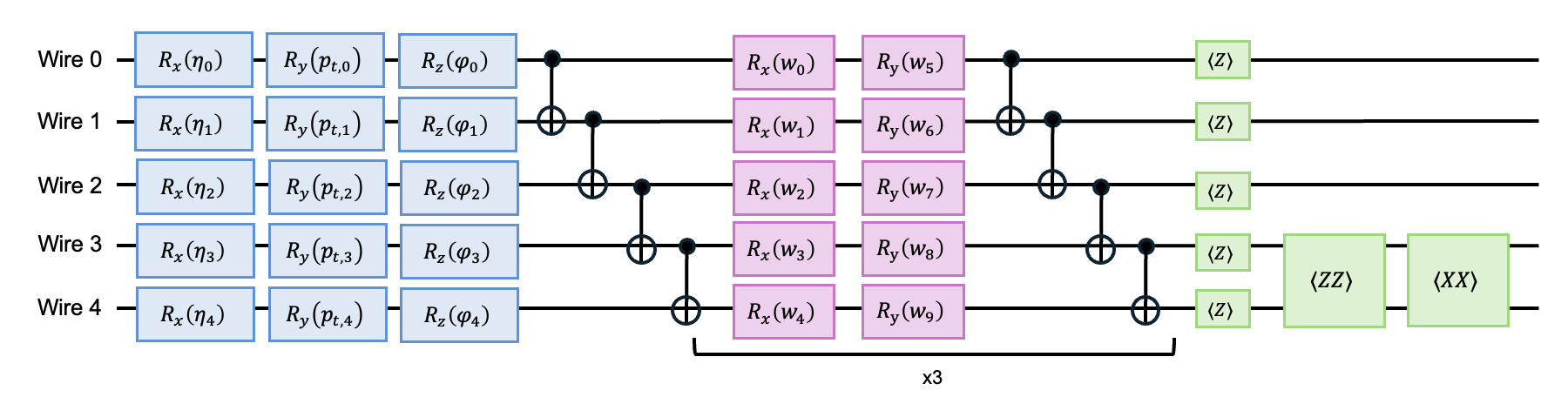} 
\caption{Architecture of one parameterized quantum circuit in the quantum autoencoder with 10 trainable parameters.% ($w_0$ to $w_9$ and $\delta$) 
\label{fig:arch_hybrid_pqc}} 
\end{figure}

Following Ref.~\cite{Govorkova_2022}, the anomaly score is taken to be the clipped Kullback-Leibler divergence (CKL) score, \begin{equation} \mathcal{S}_{\mathrm{CKL}} = \frac{1}{3}\sum_i z_{\mathrm{mean},i}^2 . \end{equation} The deployed inference encoder contains 87 trainable quantum parameters and 630 classical parameters, for 717 total trainable parameters. The decoder is used only during training and is not included in the FPGA-deployed inference path. 
The architecture is shown in Fig.~\ref{fig:arch_hybrid}, with an illustration of the PQC in Fig.~\ref{fig:arch_hybrid_pqc}. 

\subsubsection{Quantum Autoencoder} 
\begin{figure}
\centering 
\includegraphics[width=0.9\textwidth]{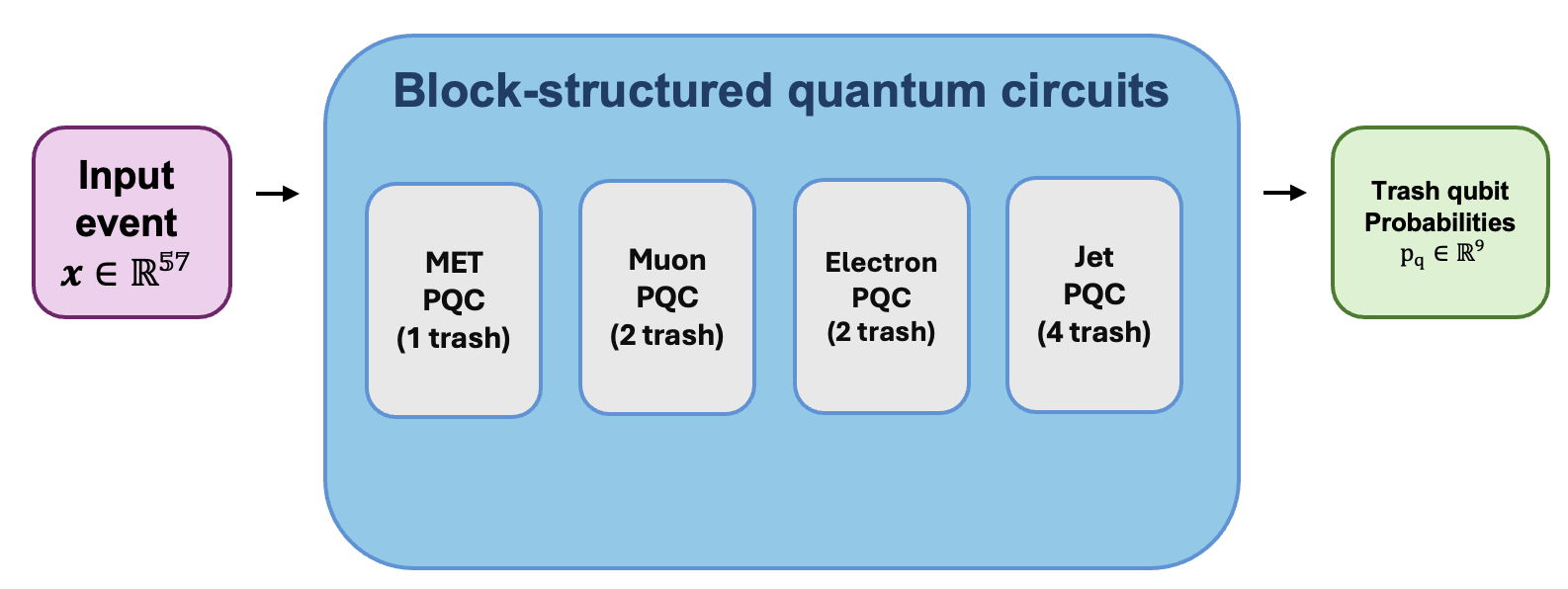} \caption{Architecture of the fully quantum autoencoder. Each particle-type sub-block is encoded, processed by trainable rotation and CRX entangling layers, and scored using background-normalized trash-wire excitation measurements. \label{fig:arch_qae}}
\end{figure}

\begin{figure} 
\centering 
\includegraphics[width=0.95\textwidth]{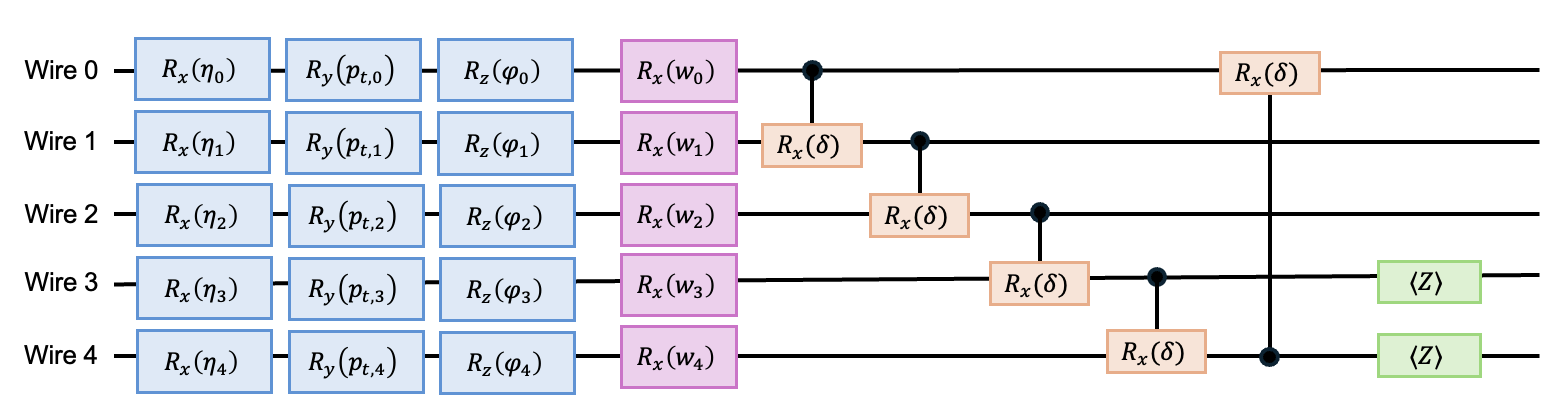} 
\caption{Architecture of one parameterized quantum circuit in the quantum autoencoder. In this PQC, there are 6 trainable parameters ($w_0$ to $w_5$ and $\delta$).  \label{fig:arch_qae_pqc}} 
\end{figure}

We also study a fully quantum autoencoder (QAE) model that removes the classical layers entirely and scores anomalies directly from trash-wire measurements.  The model uses the same 1-4-4-5-5 particle-type block partition as the hybrid VAE. Within each multi-wire block, particle features are encoded through $R_X(\eta)\,R_Y(p_\mathrm{T})\,R_Z(\phi)$ rotations on each wire. 
The circuit then applies one layer consisting of trainable RX rotations on each wire followed by a CRX ring entangler with angle $\pi/2+\delta_{\mathrm{block}}$, where $\delta_{\mathrm{block}}$ is a trainable scalar specific to each multi-wire block. 
The single-qubit $E_\mathrm{T}^\mathrm{miss}$ block does not include an entangling operation. 
The QAE objective encourages background events to be compressed into the latent qubits of each block by pushing the designated trash wires toward $|0\rangle$. 
%\tcg{SA: This next statement is unnecessarily complex.}
Each event yields nine trash measurements: one from the $E_\mathrm{T}^\mathrm{miss}$ block, two from each lepton block, and two from each jet sub-block. 
At inference time, each trash-wire excitation probability $p_i$ is converted into a background-normalized z-score as follows: \begin{equation} 
 p_i = \frac{1-\langle Z_i\rangle}{2}; \quad   z_i = \frac{p_i-\mu_i}{\sigma_i}, 
\end{equation} 
where $(\mu_i,\sigma_i)$ are fitted values. The final anomaly score is defined as the sum of the two largest positive trash-wire z-scores~\cite{ludmir2025quorumzerotrainingunsupervisedanomaly}. This model contains 23 trainable quantum parameters, consisting of 20 single-qubit rotation angles and three trainable entangler offsets $\delta_{\mathrm{block}}$, and contains no classical trainable parameters. 
The full architecture is shown in Fig.~\ref{fig:arch_qae}, with an example PQC in Fig.~\ref{fig:arch_qae_pqc}. 

\subsection{Training}\label{sec:training}

The background dataset contains 4M simulated QCD multijet events, all of which are loaded with the embedding-map normalization of Sec.~\ref{sec:embedding} prior to training. We use half of the data for training and validation (split 90/10 for train/validation), and the other half for testing. The four BSM signal files are used in their entirety only at evaluation time. 
Both models are implemented in PennyLane~\cite{bergholm2022pennylaneautomaticdifferentiationhybrid}
with PyTorch automatic differentiation and trained on background-only data
using the Adam optimizer~\cite{kingma2017adammethodstochasticoptimization}
on a single NVIDIA A100 GPU.

The hybrid VAE is trained with a batch size 512, learning rate $10^{-4}$, epoch cap
100, and early-stopping patience 15. The objective is the standard ELBO
$\mathcal{L} = \mathcal{L}_{\mathrm{recon}} + \beta\,\mathcal{L}_{\mathrm{KL}}$,
with $\beta$ cyclically annealed between 0.1 and 0.8 over 10 epochs, following~\cite{Govorkova_2022}.
The QAE is trained with a batch size of 1024 and initial learning rate $10^{-3}$, which is halved
by \texttt{ReduceLROnPlateau} after two epochs without
improvement in the validation loss. 
Early stopping is applied after 5 such epochs, with a total epoch cap of 25. 
The objective is the
background-only $z$-loss
$\mathcal{L}_z = \frac{1}{N_{\mathrm{trash}}}\sum_i (1-\langle Z_i\rangle)/2$,
driving trash excitation probabilities $p_i = P(|1\rangle_i)$ toward
zero on background events.

%--------------------------------------------------------------------------------------------------------------
\section{Results}

%--------
\subsection{Anomaly Detection Performance} 

The ability of the models to deliver anomaly detection capabilities is gauged through their ability to discriminate a variety of different signals models from the Standard Model background. 
This capacity is quantified by the receiver operating characteristic (ROC), specifically considering the area-under-curve (AUC).
While AUC is a useful quantity to provide a general sense of signal-background discrimination from a given anomaly score, it does not probe the high background rejection tails of the classifier, which is often the most relevant regime for collider analysis applications. 
Therefore the true positive rate (TPR) at a fixed background efficiency of 10$^{-5}$ is also considered. 

\begin{figure}[!tbhp]
\centering
\includegraphics[width=0.49\textwidth]{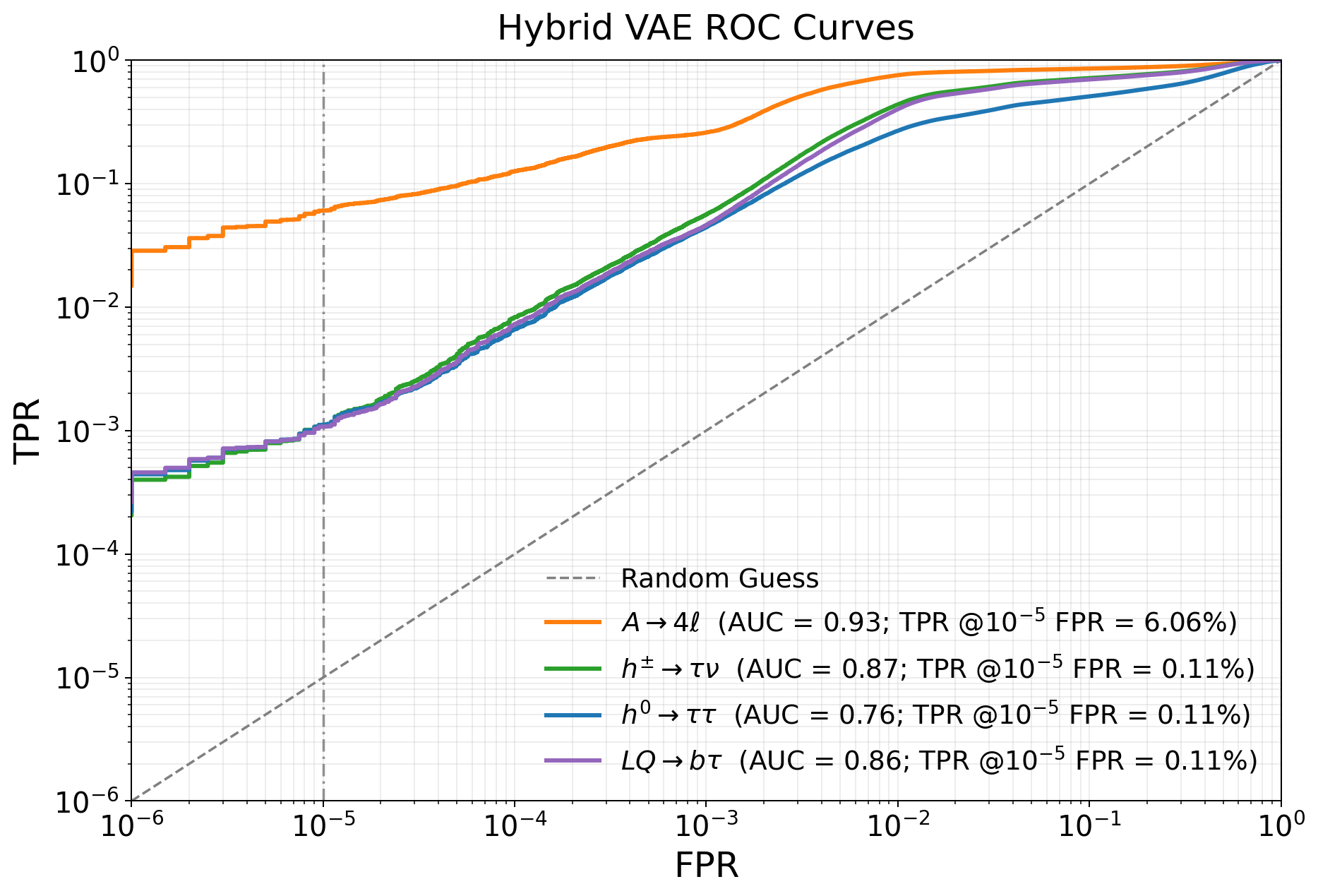}
\includegraphics[width=0.49\textwidth]{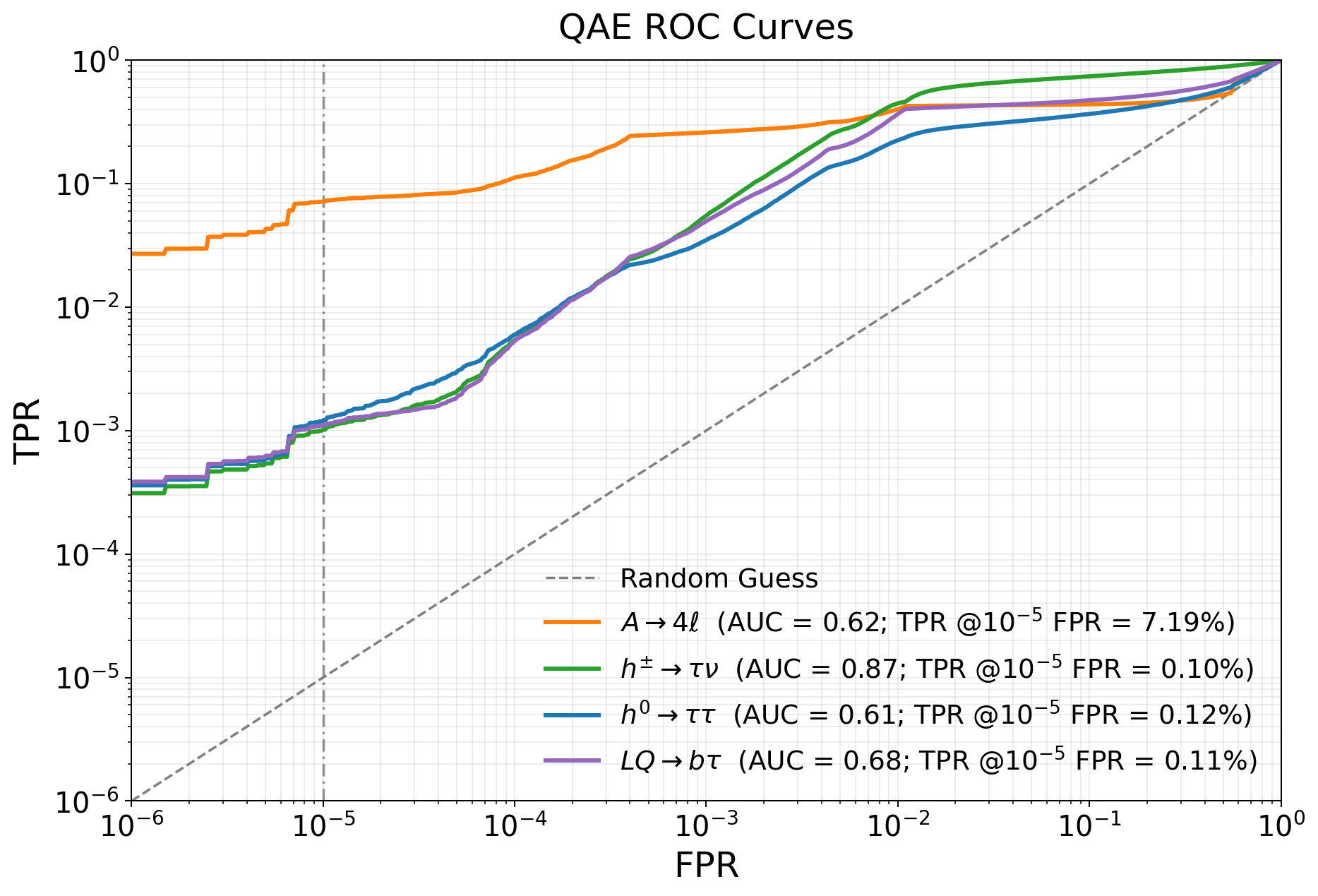}
\caption{ROC curves for the hybrid VAE (left) and QAE (right) for all four BSM signals. The red dashed line indicates the fixed false positive rate (FPR) of 10$^{-5}$ used to assess the model performance at low background acceptance for the HEP context. 
\label{fig:rocs}}
\end{figure}

Figure~\ref{fig:rocs} presents the ROC curves for both the hybrid VAE and QAE models, including all BSM signals. 
Both the AUC and TPR metrics are competitive with or exceeding performance of state-of-the-art classical ML methods~\cite{govorkova2021lhcphysicsdatasetunsupervised}.
The QAE exhibits lower AUC than the hybrid VAE for some signals. This is expected, as its training objective and score construction are designed to maximize sensitivity in the low-FPR tail relevant for triggering rather than global rank ordering. Consistent with this, its $\mathrm{TPR}@10^{-5}$ remains competitive across all four signals.

%--------
\subsection{Quantization and FPGA Synthesis}

Quantization is a useful tool to compress ML models for FPGA deployment, reducing memory footprint and arithmetic complexity while preserving sufficient accuracy to meet strict latency and resource constraints.
Both model kernels are synthesized for the Alveo~U200 FPGA at a 5.5\,ns target clock
with a 0.2\,ns tolerance. Inputs enter as IEEE-754 floats and are
cast to the encoder type at load, while weights are embedded as static
\texttt{const} ROMs. The deployed formats are summarized in
Table~\ref{tab:fixedpoint}.

\begin{table}[t]
\centering
\begin{minipage}{\linewidth}
\centering
\begin{tabular}{llll}
\hline
\textbf{Type} & \textbf{Symbol} & \textbf{Format} & \textbf{Contents} \\
\hline
\multicolumn{4}{l}{\textit{Hybrid VAE}} \\
Encoder     & $q_{\text{ae}}$ & $\langle 10,3\rangle$
  & Statevector amplitudes, $\cos/\sin$ LUTs, \\
  & & & input angles, per-wire excitations \\
Multiplier  & $\mu$ & $\langle 18,6\rangle$
  & Chained products in gate primitives \\
  & & & and $z^2$ accumulation \\
MLP acc.    & $m$ & $\langle 20,8\rangle$
  & Dense-layer accumulations \\
  & & & ($32{\to}16$, $16{\to}3$) \\
Latent      & $z$ & $\langle 18,4\rangle$
  & 3-dim Gaussian latent mean \\
Score       & $s$ & $\langle 20,5\rangle$
  & Anomaly score $\mathcal{S}_{\mathrm{CKL}}\in[0,16]$ \\
\hline
\multicolumn{4}{l}{\textit{QAE}} \\
Encoder     & $q_{\text{ae}}$ & $\langle 14,3\rangle$
  & Statevector amplitudes, $\cos/\sin$ LUTs, \\
  & & & input angles, excitations $p_k$ \\
Multiplier  & $\mu$ & $\langle 18,6\rangle$
  & Chained products in gate primitives \\
Score       & $s$ & $\langle 32,15\rangle$
  & Per-wire $z$-scores and anomaly score \\
\hline
\end{tabular}
\end{minipage}
\caption{Deployed fixed-point formats for the hybrid VAE and QAE kernels.
Formats are given as $\mathtt{ap\_fixed}\langle W, I\rangle$, with total
width $W$ and integer bits $I$.}
\label{tab:fixedpoint}
\end{table}

\begin{figure} 
\centering 
\includegraphics[width=0.49\textwidth]{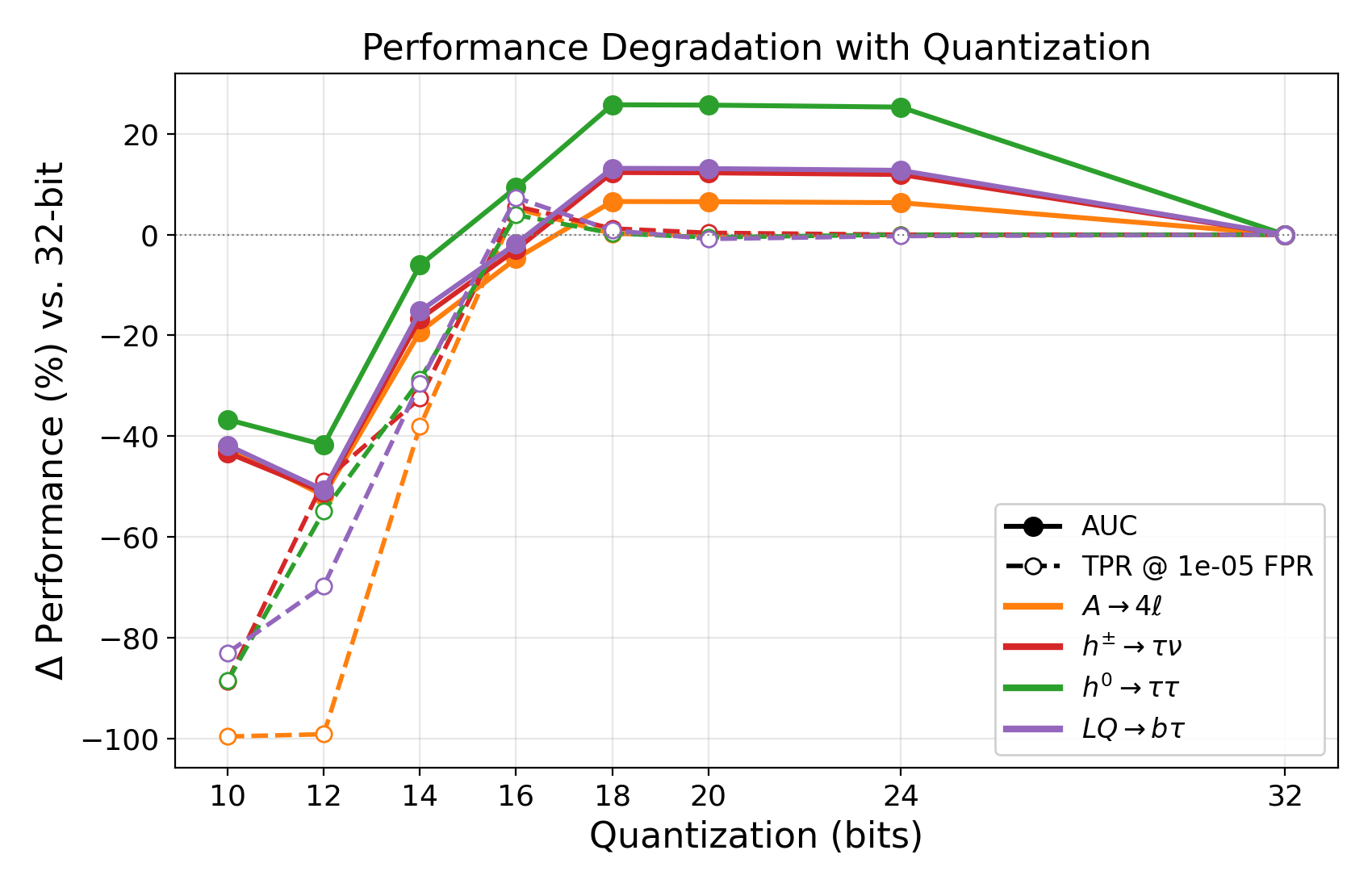} 
\includegraphics[width=0.49\textwidth]{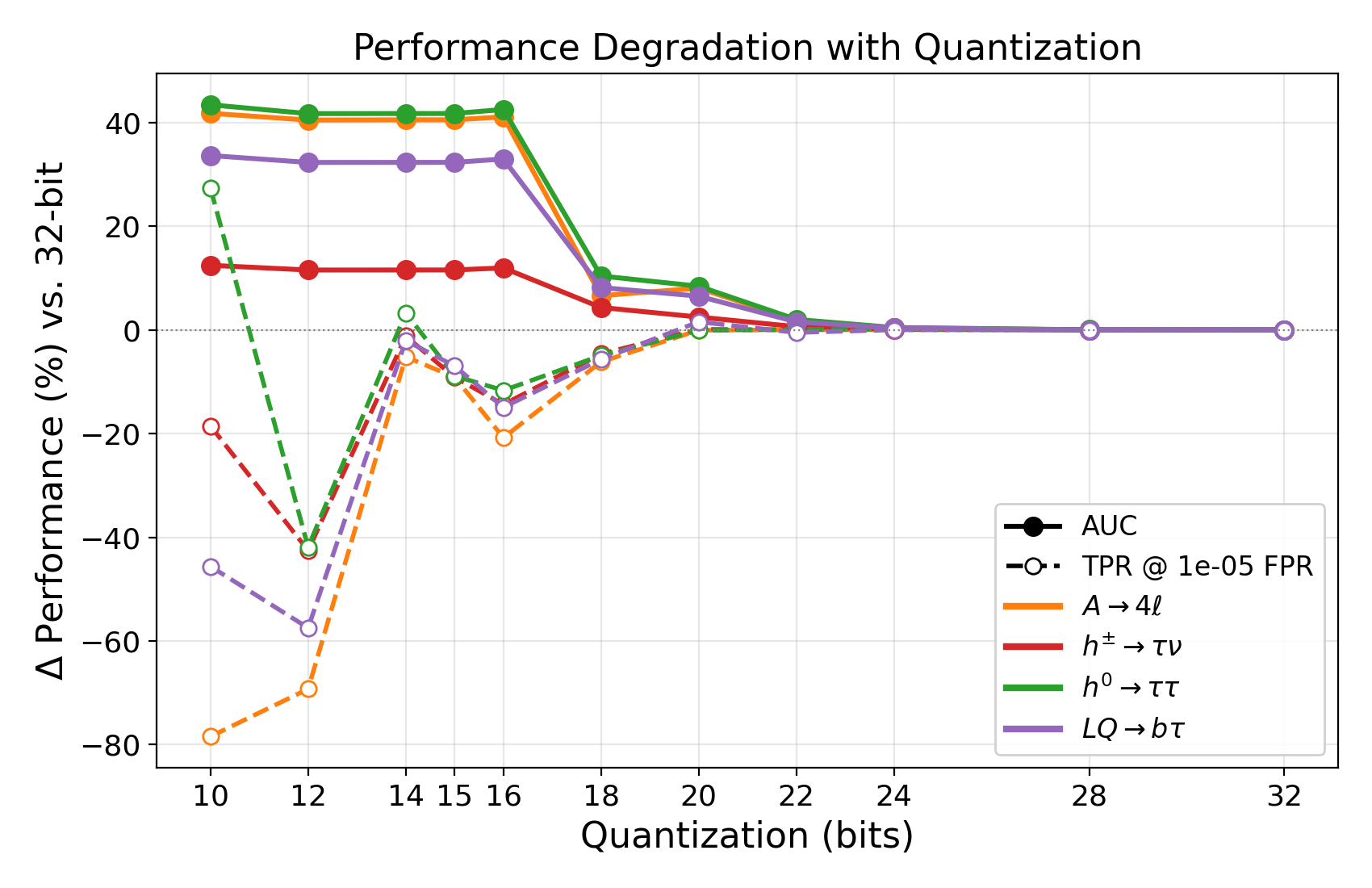}
\caption{Fixed-point quantization sweep for the Hybrid VAE (left) and the QAE (right). The x-axis shows the swept total bit-width $W$ (the first template parameter of the HLS fixed-point type) applied to each quantized datatype, while integer-bit allocations remain fixed. The curves report the relative change in AUC and $\mathrm{TPR}@10^{-5}$ FPR with respect to the 32-bit implementation.}
\label{fig:hvae_qae_quant}
\end{figure}

For the hybrid VAE, we jointly sweep the fixed-point widths $(q_{\mathrm{ae}},\mu,m,z,s)$ over total width $W\in\{10,\ldots,32\}$ as shown in Fig.~\ref{fig:hvae_qae_quant} (left). Performance collapses for $W\leq14$ due to overflow in the encoder accumulation and subsequent saturation through the re-upload chain. For $W\geq18$, the AUC becomes artificially inflated as saturation compresses the background-score tail, whereas the operating-point metric $\mathrm{TPR}@10^{-5}$ remains within $\pm5\%$ of the 32-bit baseline.

The same sweep is performed for the QAE over $(q_{\mathrm{ae}},\mu,s)$ as shown in Fig.~\ref{fig:hvae_qae_quant} (right). At lower precisions, the score distribution becomes heavily quantized, causing many background events to share identical scores. This leaves the overall anomaly-score ranking, and hence the AUC, largely unchanged or even artificially improved, while substantially perturbing the extreme upper tail that determines $\mathrm{TPR}@10^{-5}$. Consequently, the operating-point metric is a more faithful indicator of deployment performance than AUC in the low-precision regime. 

%\subsection{Agentic HLS Optimization}
Claude Code's agentic AI tools are used to optimize the FPGA resource usage for this study. A global objective of minimizing latency is set under the constraint of keeping resource consumption within a single FPGA Super Logic Region (SLR) along with a sanity check of avoiding I/O timing mismatches in parallelized algorithmic loops. The outer loop was run under the global objective until the agent found a satisfactory solution, which was used as the implementation shown in this work.

%--------
\subsection{Resource and Latency Analysis} 

Table~\ref{tab:synthesis} presents the resources and latencies of both the hybrid and quantum autoencoder models after FPGA synthesis. 
Resources are presented in terms of FPGA look-up tables (LUTs), digital signal processors (DSPs), and flip-flops (FFs). 
Both models can run in under 10 $\mu$s within the resource constraints of a single SLR, with the QAE running in under 1 $\mu$s. 
Exact final resource and latency usage is flexible and can be dictated by task-specific optimization. 
These results indicate an FPGA deployment of both the hybrid VAE and QAE that is consistent with real-time constraints of future collider trigger systems.

\begin{table}[t]
\centering
\begin{minipage}{\linewidth}
\centering
\begin{tabular}{lcc}
\hline
 & \textbf{Hybrid VAE} & \textbf{QAE} \\
\hline
\textbf{Latency} &  6.1$\: \mu$s %1109 cycles * 5.5 ns 
                 &  0.47$\: \mu$s  % 85 * 5.5 ns 
                 \\
\textbf{LUTs}    &  208976 &  249777 \\
\textbf{DSPs}    &  1528   & 1234 \\
\textbf{FFs}     &  48043 & 79429 \\
\hline
\end{tabular}
\end{minipage}
\caption{Synthesis results comparison of the hybrid VAE and QAE, considering latency and FPGA resources.}
\label{tab:synthesis}
\end{table}

\begin{figure}
\centering
\includegraphics[width=0.7\textwidth]{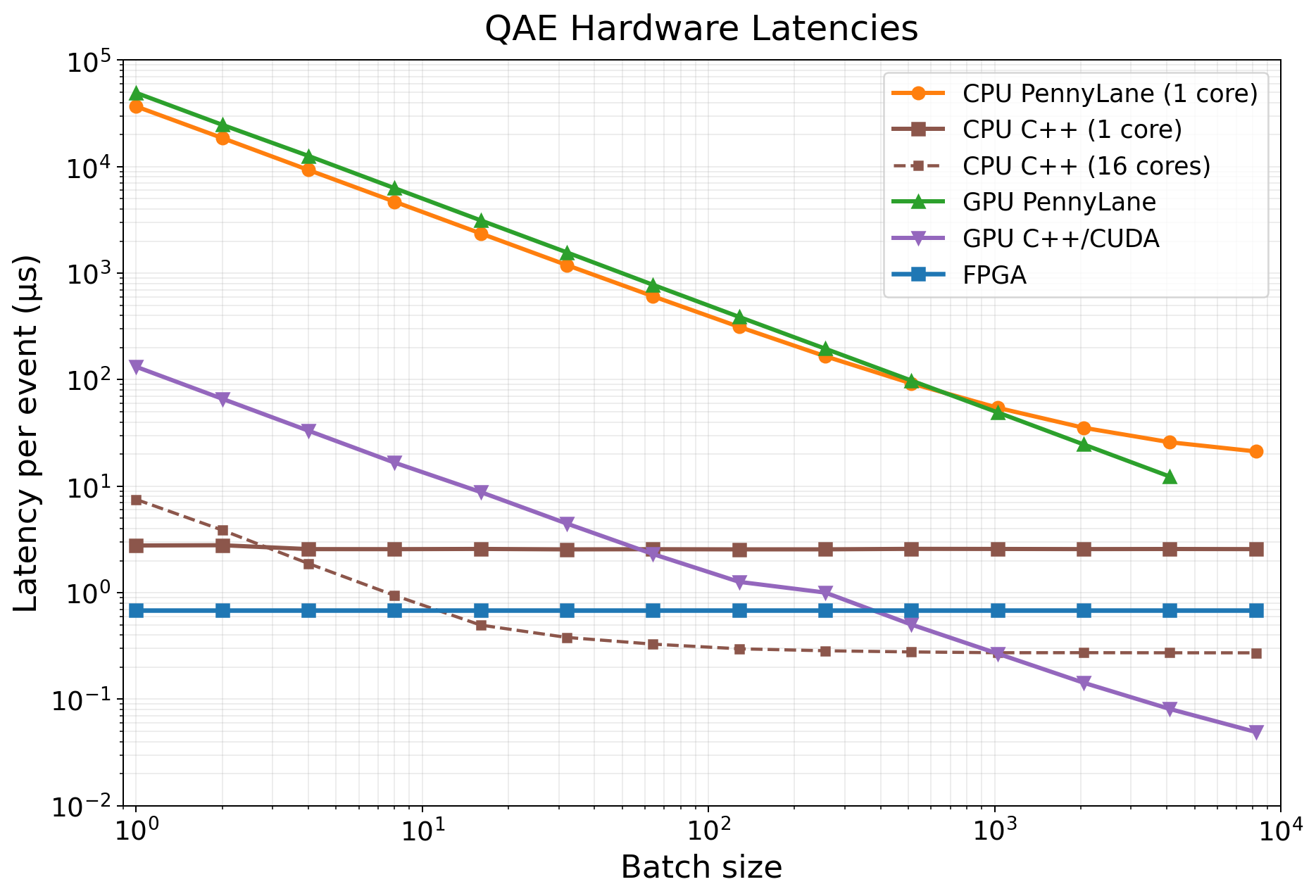}
\caption{Per-event inference latency of the QAE as a function of batch
size across hardware platforms: PennyLane state-vector simulation on a
single CPU core and on GPU, a specialized compiled C++ implementation on one and 16 CPU cores, a CUDA implementation on
GPU, and the synthesized FPGA kernel. CPU and GPU per-event latencies
fall with batch size as overheads are amortized. The trigger operating point corresponds to the left edge
of the plot.
\label{fig:hw_comparison}}
\end{figure}

Furthermore, we compare per-event QAE inference latency across CPU, GPU, and FPGA implementations as a function of batch size in Figure~\ref{fig:hw_comparison}.
At a batch size of one, which is the operating regime of the trigger system, the FPGA kernel is the fastest
option. However, as batch size is increased, multi-threaded CPU and CUDA implementations surpass the FPGA in
per-event throughput. This behavior is consistent with prior observations that FPGA emulation is advantaged for small quantum registers where per-event work is too limited to occupy massively parallel hardware~\cite{10.1109/ISVLSI.2008.43, vanduy2025hpqeascalablehighperformancequantum}.

%--------------------------------------------------------------------------------------------------------------
\section{Conclusions}

We present quantum and classical-quantum hybrid autoencoder models developed for real-time anomaly detection in future collider experiments. 
Both models leverage rotation gates to embed classical event features into per-particle quantum states and entangle groups of particles together.
The resulting models can meet or exceed state-of-the-art anomaly detection capability, as measured by breadth of signal-background discrimination, using fewer than 100 trainable quantum parameters. 
Furthermore, after quantization and FPGA synthesis, both models can meet latency requirements of future colliders and fit within 1 FPGA SLR, with the QAE final latency being below 1$\mu$s. 
These results indicate the potential for quantum ML, deployed via classical FPGA emulation, to complement classical methods for real-time inference in detector readout. 
Future work could include an agentic AI approach to optimizing the algorithmic implementation of the quantum gates in synergy with the HLS statements, and a hybrid quantum gate-tensor network implementation tailored to FPGA resource minimization at performance parity.

\backmatter

%%%%%%%%%%%%%%%%%%%%%%%%%%%%%%%%%%%%%%%%%%%%%%%%%%%%%%%%%%%%%%%%%%%%%%%%%%%%
\section*{Funding}

This work is supported by the U.S. Department of Energy under contract number DE-AC02-76SF00515 and the Office of the Vice Provost for Undergraduate Education at Stanford University.

%%%%%%%%%%%%%%%%%%%%%%%%%%%%%%%%%%%%%%%%%%%%%%%%%%%%%%%%%%%%%%%%%%%%%%%%%%%%
\section*{Author Contributions}

IG was the primary developer of the ML models and methods, produced all final results and figures, and contributed to paper writing. SA and AD provided expert consultation on the QML and hardware engineering, respectively, required to produce the results. JG supervised the work, and contributed to writing and editing the paper. 

%%%%%%%%%%%%%%%%%%%%%%%%%%%%%%%%%%%%%%%%%%%%%%%%%%%%%%%%%%%%%%%%%%%%%%%%%%%%
\section*{Competing Interests}

The authors declare no competing interests.

%%%%%%%%%%%%%%%%%%%%%%%%%%%%%%%%%%%%%%%%%%%%%%%%%%%%%%%%%%%%%%%%%%%%%%%%%%%%
\section*{Data Availability}

The background and signal samples used in this study are available on Zenodo~\cite{thea_aarrestad_2021_5046389,thea_aarrestad_2021_5046446,thea_aarrestad_2021_5061633,thea_aarrestad_2021_7152617,thea_aarrestad_2021_5055454}.

%%%%%%%%%%%%%%%%%%%%%%%%%%%%%%%%%%%%%%%%%%%%%%%%%%%%%%%%%%%%%%%%%%%%%%%%%%%%
\section*{Code Availability}
The relevant machine learning code and the high-level synthesis implementation are publicly available at \url{https://github.com/SLAC-Julia-Group/hardware-aware-quantum-autoencoders}.

\bibliography{qae_fpga}% common bib file

\end{document}